\RequirePackage{fix-cm}
\documentclass[twocolumn]{svjour3}          % twocolumn
\smartqed  % flush right qed marks, e.g. at end of proof
\usepackage{graphicx}
\usepackage{booktabs}
\usepackage{array}
\usepackage{multirow}
\usepackage{hyphenat}
\usepackage{url}
\usepackage[round]{natbib}
\usepackage[pagebackref=false,breaklinks=true,letterpaper=true,colorlinks,bookmarks=false]{hyperref}
\newcolumntype{C}[1]{>{\centering\arraybackslash}p{#1}}
\begin{document}

\title{Salient Object Subitizing%\thanks{Grants or other notes
}


\author{Jianming Zhang         \and
        Shugao Ma              \and
        Mehrnoosh Sameki       \and
        Stan Sclaroff          \and
        Margrit Betke          \and
        Zhe Lin                \and
        Xiaohui Shen           \and
        Brian Price            \and
        Radom\'{i}r M\v{e}ch %etc.
}


\institute{Jianming Zhang, Mehrnoosh Sameki, Stan Sclaroff, Margrit Betke \at
              Computer Science Dept., Boston Univ., Boston, MA USA\\
              \email{\{jmzhang,sameki,sclaroff,betke\}@bu.edu}           %  \\
           \and
           Shugao Ma \at Oculus Research Pittsburgh, Pittsburgh, PA USA\\
           \email{shugao.ma@oculus.com }
           \and
           Zhe Lin, Xiaohui Shen, Brian Price, Radom\'{i}r M\v{e}ch \at
              Adobe Research, San Jose, CA USA\\
              \email{\{zlin,xshen,bprice,rmech\}@adobe.com}
}

\date{Received: date / Accepted: date}

\maketitle

\begin{abstract}
We study the problem of Salient Object Subitizing, \emph{i.e.}\ predicting the existence and the number of salient objects in an image using holistic cues. This task is inspired by the ability of people to quickly and accurately identify the number of items within the subitizing range (1-4). To this end, we present a salient object subitizing image dataset of about 14K everyday images which are annotated using an online crowdsourcing marketplace. We show that using an end-to-end trained Convolutional Neural Network (CNN) model, we achieve prediction accuracy comparable to human performance in identifying images with zero or one salient object. For images with multiple salient objects, our model also provides significantly better than chance performance without requiring any localization process. Moreover, we propose a method to improve the training of the CNN subitizing model by leveraging synthetic images. In experiments, we demonstrate the accuracy and generalizability of our CNN subitizing model and its applications in salient object detection and image retrieval.

\end{abstract}

\section{Introduction}

\begin{figure}
  \centering
  \includegraphics[width=1\linewidth]{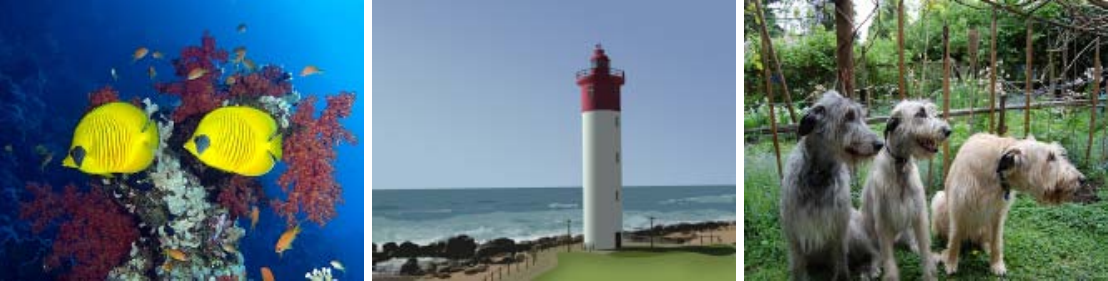}\\
  \caption{How fast can you tell the number of prominent objects in each of these images? It is easy for people to identify the number of items in the range of 1-4 by a simple glance. This ``fast counting" ability is known as \emph{Subitizing}.}\label{fig:teaser}
\end{figure}

How quickly can you tell the number of \textbf{salient} objects in each image in Fig.~\ref{fig:teaser}?

\begin{figure*}
	\includegraphics[width = 1\linewidth]{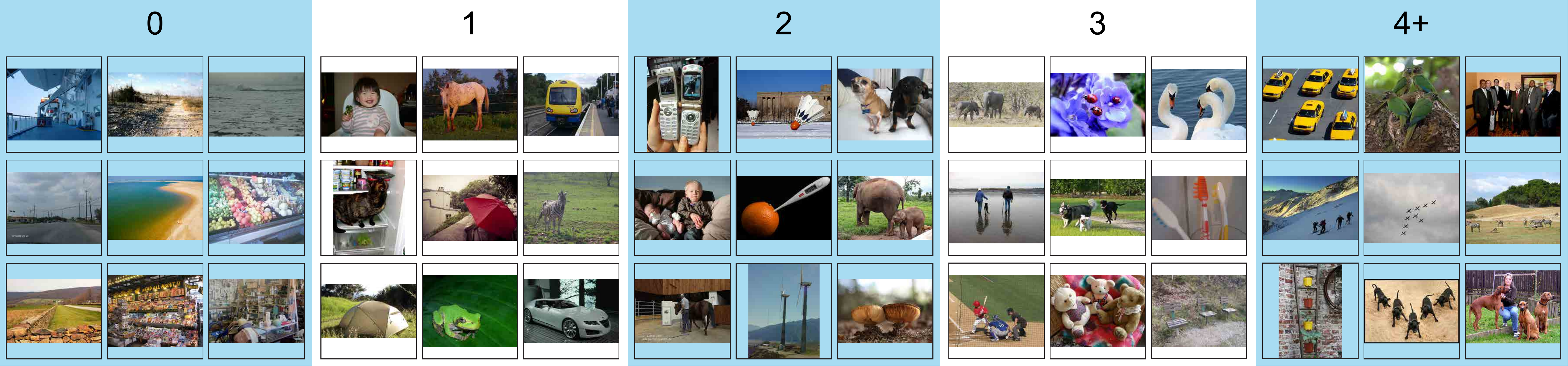}
    \caption{Sample images of the proposed SOS dataset. We collected about 14K everyday images, and use Amazon Mechanical Turk (AMT) to annotate the number of salient object of each image. The consolidated annotation is shown on the top of each image group. These images cover a wide range of content and object categories.}\label{fig:frontpage}
\end{figure*}
As early as the 19th century, it was observed that humans can effortlessly identify the number of items in the range of 1-4 by a glance \citep{jevons1871power}. Since then, this phenomenon, later coined by Kaufman \emph{et al.}\ as \textit{Subitizing} \citep{kaufman1949discrimination}, has been studied and tested in various experimental settings \citep{atkinson1976magic, mandler1982subitizing}. It is shown that identifying small numbers up to three or four is highly accurate, quick and confident, while beyond this subitizing range, this sense is lost. Accumulating evidence also shows that infants and even certain species of animals can differentiate between small numbers of items within the subitizing range \citep{dehaene2011number,gross2009number,davis1988numerical,pahl2013numerical}. This suggests that subitizing may be an inborn numeric capacity of humans and animals. It is speculated that subitizing is a preattentive and  parallel process \citep{dehaene2011number,trick1994small,vuilleumier2000systematic}, and that it can help humans and animals make prompt decisions in basic tasks like navigation, searching and choice making \citep{piazza2004number,gross2012magical}.

Inspired by the subitizing phenomenon, we propose to study the problem of \textit{Salient Object Subitizing} (SOS), \emph{i.e.}\ predicting the existence and the number (1, 2, 3, and 4+) of salient objects in an image without using any localization process. Solving the SOS problem can benefit many computer vision tasks and applications.

Knowing the existence and the number of salient objects without the expensive detection process can enable a machine vision system to select different processing pipelines at an early stage, making it more intelligent and reducing computational cost. For example,  SOS can help a machine vision system suppress the object recognition process, until the existence of salient objects is detected, and it can also provide cues for generating a proper number of salient object detection windows for subsequent processing. Furthermore, differentiating between scenes with zero, a single and multiple salient objects can also facilitate applications like image retrieval, iconic image detection \citep{berg2009finding}, image thumbnailing \citep{choi2014determining}, robot vision \citep{scharfenberger2013existence}, egocentric video summarization \citep{lee2012discovering}, snap point prediction \citep{xiong2014detecting}, \emph{etc}.

In our preliminary work \citep{zhang2015salient}, we presented the first formulation of SOS and an SOS image dataset of about 7K images. The number of salient objects in each image was annotated by Amazon Mechanical Turk (AMT) workers. The resulting annotations from the AMT workers were analyzed in a more controlled offline setting; this analysis showed a high inter-subject consistency in subitizing salient objects in the collected images.
In this paper, we follow the same data collection procedure and expand our SOS dataset by approximately doubling the dataset size. This allows us to train more generalizable SOS models and have more robust evaluations. In Fig.~\ref{fig:frontpage}, we show some sample images in the SOS dataset with the collected groundtruth labels.
%Although there are no bounding box annotations accompanying the numbers, it is usually pretty straightforward to see which objects these numbers refer to.

We formulate the SOS problem as an image classification task, and aim to develop a method to quickly and accurately predict the existence and the number of generic salient objects in everyday images. We propose to use an end-to-end trained Convolutional Neural Network (CNN) model for our task, and show that an implementation of our method achieves very promising performance. In particular, the CNN-based subitizing model can approach human performance in identifying images with no salient object and with a single salient object. We visualize the learned CNN features and show that these features are quite generic and discriminative for the class-agnostic task of subitizing. Moreover, we empirically validate the generalizability of the CNN subitizing model to unseen object categories.

To further improve the training of the CNN SOS model, we experiment with the usage of synthetic images. We generate a total of 20K synthetic images that contain different numbers of dominant objects using segmented objects and background images. We show that model pre-training using these synthetic images results in an absolute increase of more than 2\% in Average Precision (AP) in identifying images with 2, 3 and 4+ salient objects respectively. In particular, for images with 3 salient objects, our CNN model attains an absolute increase of about 6\% in AP.

We demonstrate the application of our SOS method in salient object detection and image retrieval. For salient object detection, our SOS model can effectively suppress false object detections on background images and estimate a proper number of detections. By leveraging the SOS model, we attain an absolute increase of about 4\% in F-measure over the state-of-the-art performance in unconstrained salient detection \citep{zhang2015SOD}. For image retrieval, we show that the SOS method can be used to handle queries with object number constraints.

In summary, the key contributions of this work are:
    \begin{enumerate}
        \item We formulate the Salient Object Subitizing (SOS) problem, which aims to predict the number of salient objects in an image without resorting to any object localization process.
        \item We provide a large-scale image dataset for studying the SOS problem and benchmarking SOS models.
% Old:       \item We present a CNN-based implementation of the proposed SOS method, which achieves promising results, while being capable of processing an image in milliseconds.
%new:
         \item We present a CNN-based method for SOS, and propose to use synthetic images to improve the learned CNN model.
         \item We demonstrate applications of the SOS method in salient object detection and image retrieval.
    \end{enumerate}

Compared with our preliminary work on SOS \citep{zhang2015salient}, we make several major improvements in this paper: 1) we expand the SOS dataset by doubling the number of images; 2) we attains significantly better performance by leveraging a more advanced CNN architecture, additional real training data and a large number of synthetic training data; 3) we conduct extensive experimental analyses to compare CNN model architectures, visualize the learned CNN features, and validate the generalizability of the SOS model for unseen object categories; 4) in addition to salient object detection, we demonstrate the application of SOS in image retrieval.

\section{Related Work}

%We review the following areas that are related to our task.

{\bf Salient object detection.} Salient object detection aims at detecting dominant objects in a scene. Given a test image, some methods \citep{achanta2009frequency,cheng2011global,shen2012unified,zhang2015MBD} generate a saliency map that highlights the overall region of salient objects; other methods \citep{liu2011learning,gopalakrishnan2009random,feng2011salient,siva2013looking} produce bounding boxes for localization. Ideally, if a salient object detection method can well localize each salient object, then the number of objects can be simply inferred by counting the detection windows. However, many existing salient object detection methods assume the existence of salient objects, and they are mainly tested and optimized for images that contain a single dominant object \citep{li2014secrets,borji2012salient}. Therefore, salient object detection methods often generate undesirable results on background images, and are prone to fail on images with multiple objects and complex background. Recently, \cite{zhang2015SOD} proposed a salient object detection method for unconstrained images. Although this method can handle complex images to some extent, we will show that the counting-by-detection approach is less effective than our subitizing method in predicting the number of salient objects.

{\bf Detecting the existence of salient objects.} Only a few works address the problem of detecting the existence of salient objects in an image. \cite{wang2012salient} use a global feature based on several saliency maps to determine the existence of salient objects in thumbnail images. Their method assumes that an image either contains a single salient object or none. \cite{scharfenberger2013existence} use saliency histogram features to detect the existence of interesting objects for robot vision. It is worth noting that the testing images handled by the methods of \cite{wang2012salient} and \cite{scharfenberger2013existence} are substantially simplified compared to ours, and these methods cannot predict the number of salient objects.
%In \cite{choi2014determining},  the existence of salient object is inferred via a local patch approach combined with a Bayesian methodology. However,  the method of \cite{choi2014determining} is basically designed to differentiate scene images from object images. As we will show, many images from common object datasets do not contain clearly defined salient objects.

{\bf Automated object counting.} There is a large body of literature about automated object counting based on density estimation \citep{lempitsky2010learning, arteta2014interactive}, object detection/segmentation \citep{subburaman2012counting, nath2006cell,anoraganingrum1999cell} and regression \citep{chan2008privacy, chan2009bayesian}. While automated object counting methods are often designed for crowded scenes with many objects to count, the SOS problem aims to discriminate between images with 0, 1, 2, 3 and 4+ dominant objects.
Moreover, automated object counting usually focuses on a specific object category (\emph{e.g.}\ people and cells), and assumes that the target objects have similar appearances and sizes in the testing scenario. On the contrary, the SOS problem addresses category-independent inference of the number of salient objects. The appearance and size of salient objects can vary dramatically from category to category, and from image to image, which poses a very different challenge than the traditional object counting problem.

{\bf Modeling visual numerosity.} Some researchers exploit deep neural network models to analyze the emergence of visual numerosity in human and animals \citep{stoianov2012emergence,Zou:COGSCI13}. In these works, abstract binary patterns are used as training data, and the researchers study how the deep neural network model captures the number sense during either unsupervised or supervised learning. Our work looks at a more application-oriented problem, and targets at inferring the number of salient objects in natural images.

\section{The SOS Dataset}

We present the Salient Object Subitizing (SOS) dataset, which contains about 14K everyday images. This dataset expands the dataset of about 7K images reported in our preliminary work \citep{zhang2015salient}. We first describe the collection of this dataset, and then provide a human labeling consistency analysis for the collected images. The dataset is available on our project website\footnote{\url{http://www.cs.bu.edu/groups/ivc/Subitizing/}}.

\subsection{Image Source}
\begin{figure}
	\includegraphics[width = 1\linewidth]{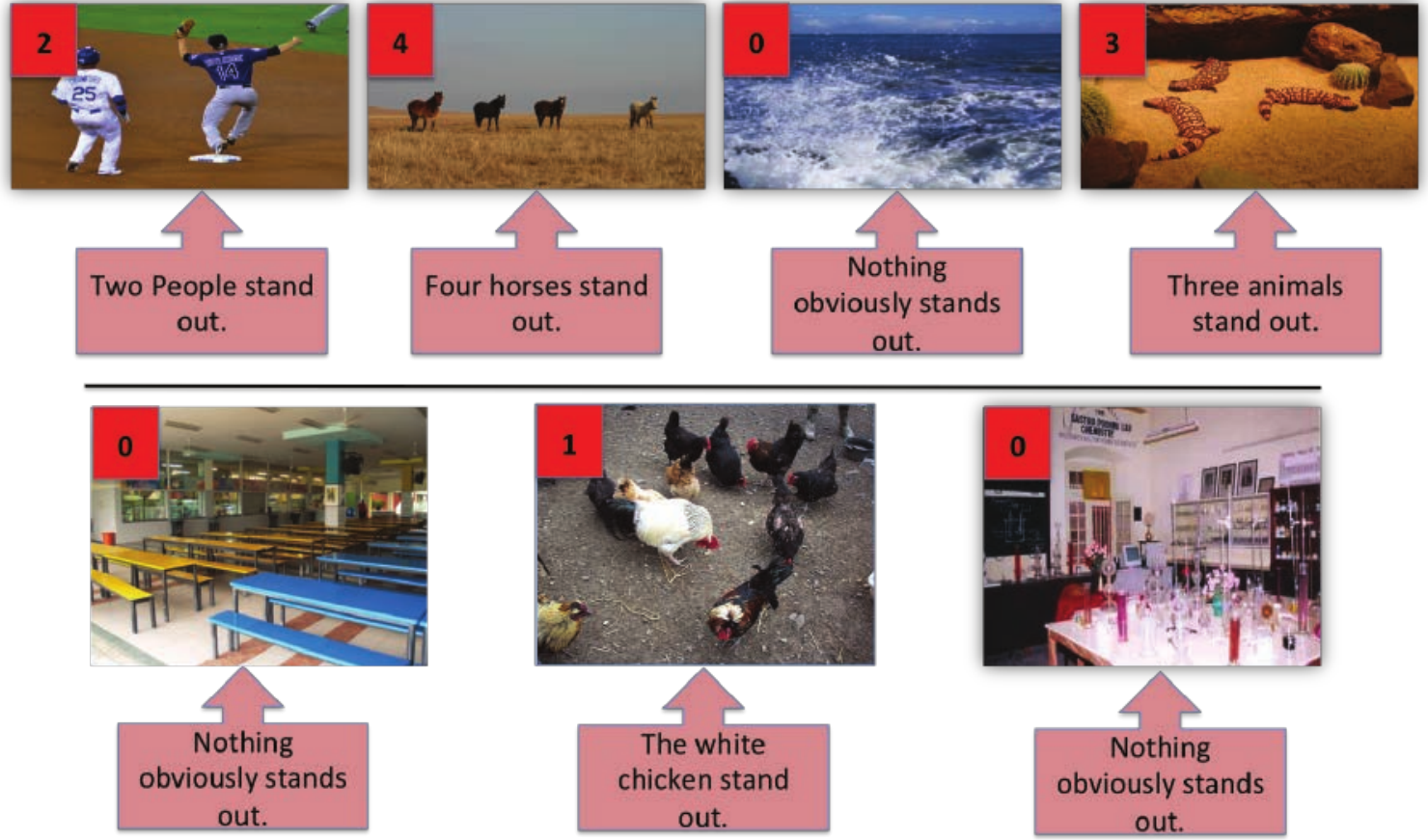}
    \caption{Example labeled images for AMT workers. The number of salient objects is shown in the red rectangle on each image. There is a brief explanation below each image.}\label{fig:MT}
\end{figure}
To collect a dataset of images with different numbers of salient objects, we gathered an initial set of images from four popular image datasets, COCO \citep{COCO}, ImageNet \citep{ILSVRCarxiv14}, VOC07 \citep{pascal-voc-2007}, and SUN \citep{xiao2010sun}. Among these datasets, COCO, ImageNet and VOC07 are designed for object detection, while SUN is for scene classification. Images from COCO and VOC07 often have complex backgrounds, but their content is limited to common objects and scenes. ImageNet contains a more diverse set of object categories, but most of its images have centered dominant objects with relatively simpler backgrounds. In the SUN dataset, many images are rather cluttered and do not contain any salient objects. We believe that combining images from different datasets can mitigate the potential data bias of each individual dataset.

This preliminary set is composed of about 30000 images in total. There are about 5000 images from SUN, 5000 images from VOC07 respectively, 10000 images are from COCO and 10000 images from ImageNet. For VOC07, the whole training and validation sets are included. We limited the number of images from the SUN dataset to 5000, because most images in this dataset do not contain obviously salient objects, and we do not want the images from this dataset to dominate the category for background images. The 5000 images were randomly sampled from SUN. For the COCO and ImageNet datasets\footnote{We use the subset of ImageNet images with bounding box annotations.}, we used the bounding box annotations to split the dataset into four categories for 1, 2, 3 and 4+, and then sampled an equal number of images from each category, in the hope that this can help balance the distribution of our final dataset.

\subsection{Annotation Collection}

We used the crowdsourcing platform Amazon Mechanical Turk (AMT) to collect annotations for our preliminary set of images. We asked the AMT workers to label each image as containing 0, 1, 2, 3 or 4+ prominent objects. Several example labeled images (shown in Fig.\ \ref{fig:MT}) were provided prior to each task as an instruction. We purposely did not give more specific instructions regarding some ambiguous cases for counting, \emph{e.g.}\ counting a man riding a horse as one or two objects. We expected that ambiguous images would lead to divergent annotations.

Each task, or HIT (Human Intelligence Task) was composed of five to ten images with a two-minute time limit, and the compensation was one to two cents per task. All the images in one task were displayed at the same time. The average completion time per image was about 4s.
% commented out by Stan
%, leading to an hourly rate of about 2 dollars.
We collected five annotations per image from distinct workers. About 800 workers contributed to this dataset. The overall cost for collecting the annotation is about 600 US dollars including the fees paid to the AMT platform.
\begin{table}
\centering
\caption{Distribution of images in the SOS dataset}\label{tab:data}
\begin{tabular}{cccccc}
	\toprule
    category 		&COCO		&VOC07		&ImageNet	&SUN		&total\\\cmidrule{1-6}
    0				&616		&311        &371		&1963		&3261\\
    1               &2504       &1691       &1516       &330        &6041\\
    2				&585        &434        &935        &76         &2030\\
    3               &244        &106        &916        &43         &1309\\
    4+              &371        &182        &475        &38         &1066\\\cmidrule{1-6}
    total           &4320       &2724       &4213       &2450       &13707\\\bottomrule
\end{tabular}
\end{table}

\begin{figure}
	\includegraphics[width = 1\linewidth]{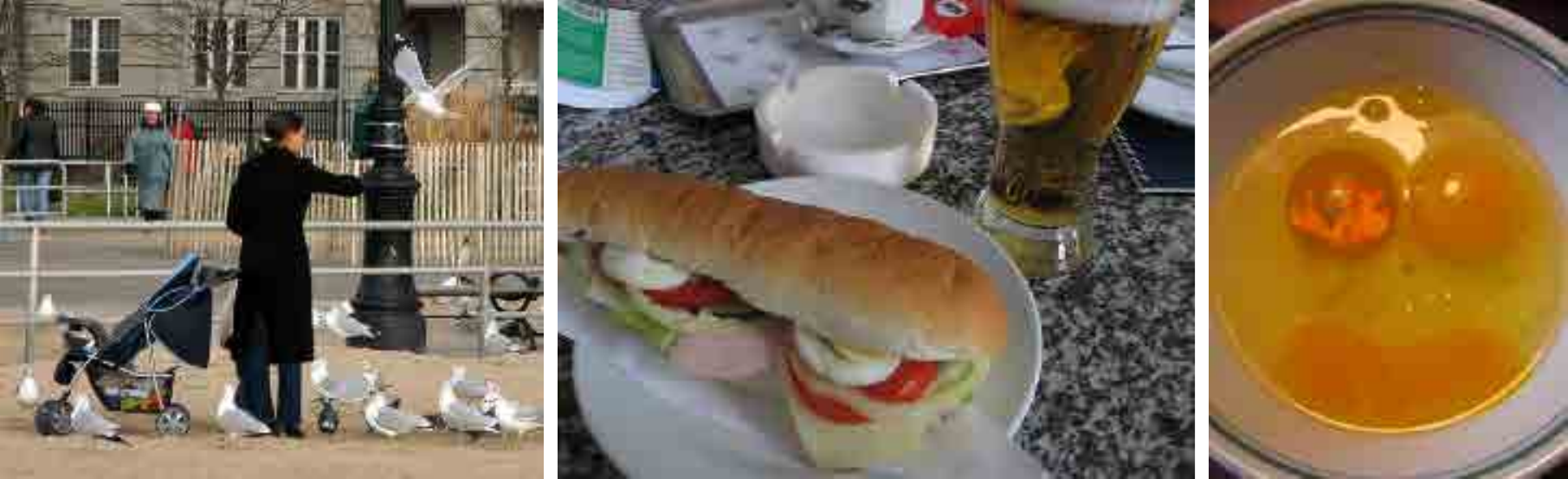}
    \caption{Sample images with divergent labels. These images are a bit ambiguous about what should be counted as an individual salient object. We exclude this type of images from the final SOS dataset. }\label{fig:div}
\end{figure}

A few images do not have a clear notion about what should be counted as an individual salient object, and labels on those images tend to be divergent. We show some of these images in Fig.~\ref{fig:div}. We exclude images with fewer than four consensus labels, leaving about 14K images for our final SOS dataset. In Table \ref{tab:data}, we show the joint distribution of images with respect to the labeled category and the original dataset. As expected, the majority of the images from the SUN dataset belong to the ``0" category. The ImageNet dataset contains significantly more images with two and three salient objects than the other datasets.

\subsection{Annotation Consistency Analysis}

During the annotation collection process, we simplified the task for the AMT workers by giving them 2 minutes to label five images at a time. This simplification allowed us to gather a large number of annotations with reduced time and cost. However, the flexible viewing time allowed the AMT workers to look closely at these images, which may have had an influence over their attention and their answers to the number of salient objects. This leaves us with a couple important questions. Given a shorter viewing time, will labeling consistency among different subjects decrease? Moreover, will shortening the viewing time change the common answers to the number of salient objects? Answering these question is critical in understanding our problem and dataset.
%Answering these questions can help us better understand our problem and data, and it will also provide a human subitizing performance baseline on our SOS dataset.

\begin{figure}
\centering
	\includegraphics[width = 0.47\linewidth]{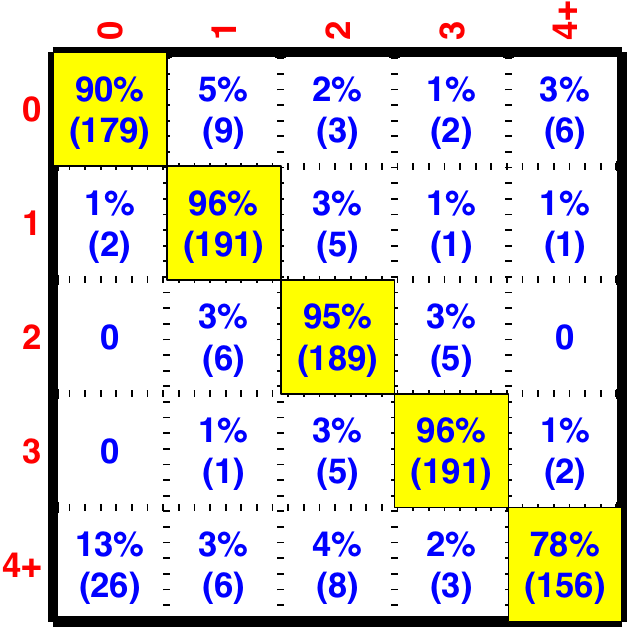}
    \caption{Averaged confusion matrix of our offline human subitizing test. Each row corresponds to a groundtruth category labeled by AMT workers. The percentage reported in each cell is the average proportion of images of the category A (row number) labeled as category B (column number). For over $90\%$ images, the labels from the offline subitizing test are consistent with the labels from AMT workers. }\label{fig:human_conf}
\end{figure}

\begin{table}
\centering
\caption{Human subitizing accuracy in matching category labels from Mechanical Turk workers.}\label{tag:human_acc}
\begin{tabular}{ccccc}
\toprule
					& sbj.1				& sbj.2				&sbj.3				& Avg.\ \\\cmidrule{1-5}
Accuracy& 90\%			  & 92\%		    &90\%			  &91\%\\\bottomrule

\end{tabular}
\end{table}

To answer these questions, we conducted a more controlled offline experiment based on common experimental settings in the subitizing literature \citep{atkinson1976magic,mandler1982subitizing}. In this experiment, only one image was shown to a subject at a time, and this image was exposed to the subject for only 500 ms. After that, the subject was asked to tell the number of salient objects by choosing an answer from 0, 1, 2, 3,  and 4+.

\begin{figure}
\centering
	\includegraphics[width = 0.9\linewidth]{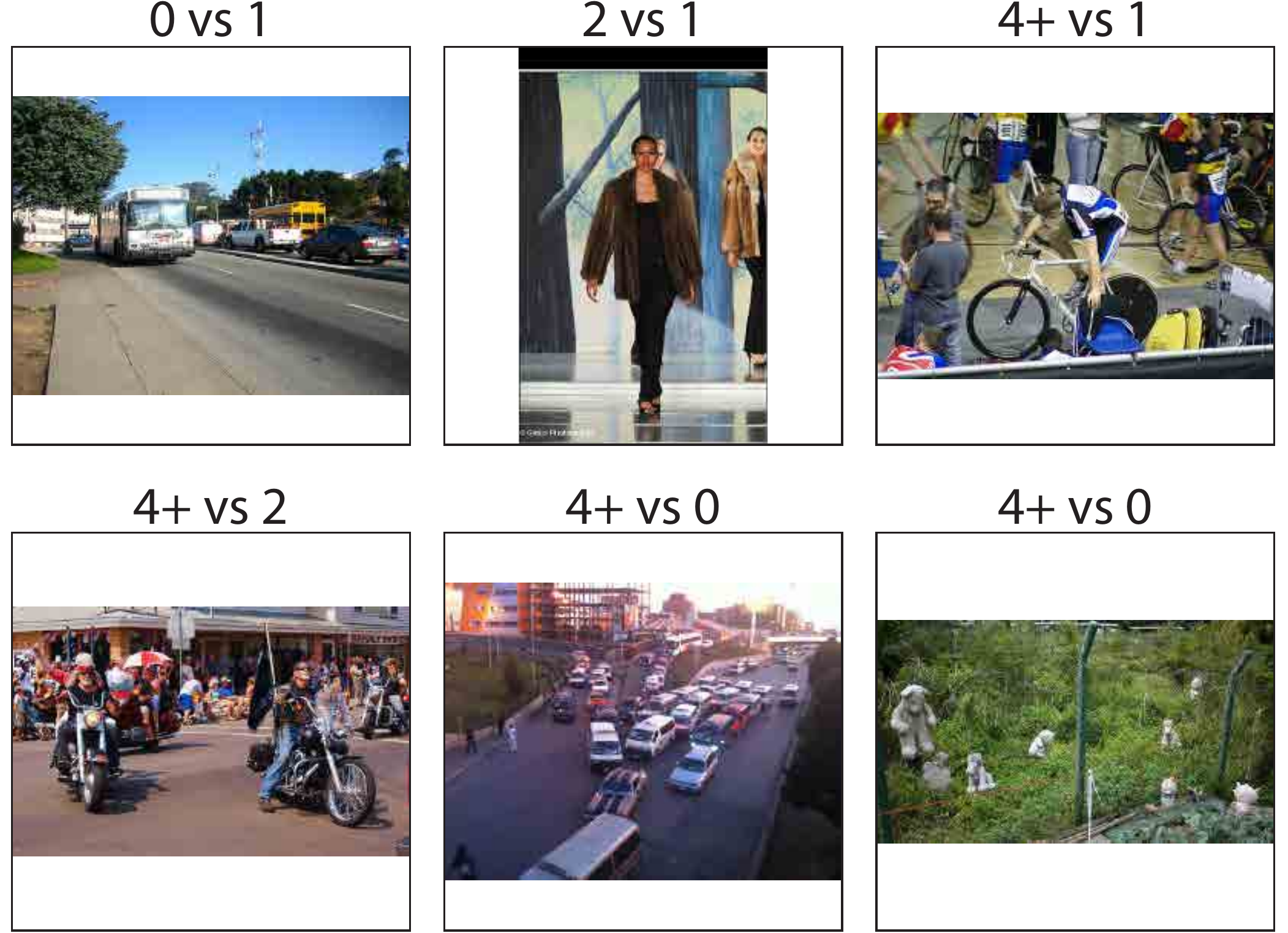}
    \caption{Sample images that are consistently labeled by all three subjects in our offline subitizing test as a different category from what is labeled by the Mechanical Turk workers. Above each image, there is the AMT workers' label (left) vs  the offline-subitizing label (right).}\label{fig:sbt_amt_diff}
\end{figure}

We randomly selected 200 images from each category according to the labels collected from AMT.
%This offline experiment was reported in our preliminary work \citep{zhang2015salient}. The images used in this offline experiment are sampled from the 6900 images of our original dataset, which is a subset of our full dataset. These sampled images can still represent our expanded dataset because the additional images are from the same sources and are annotated by crowdworkers in the same way.
Three subjects were recruited for this experiment, and each of them was asked to complete the labeling of all 1000 images. We divided that task into 40 sessions, each of which was composed of 25 images. The subjects received the same instructions as the AMT workers, except they were exposed to one image at a time for 500 ms. Again, we intentionally omitted specific instructions for ambiguous cases for counting.

Over $98\%$ test images receive at least two out of three consensus labels in our experiment, and  all three subjects agree on $84\%$ of the test images. Table~\ref{tag:human_acc} shows the proportion of category labels from each subject that match the labels from AMT workers. All subjects agree with AMT workers  on over $90\%$ of sampled images. To see details of the labeling consistency,  we show in Fig.~\ref{fig:human_conf} the averaged confusion matrix of the three subjects. Each row corresponds to a category label from the AMT workers, and in each cell, we show the average number (in the brackets) and percentage of images of category A (row number) classified as category B (column number). For categories 1, 2 and 3, the per-class accuracy scores are above $95\%$, showing that limiting the viewing time has little effect on the answers in these categories. For category 0, there is a $90\%$ agreement between the labels from AMT workers and from the offline subitizing test, indicating that changing the viewing time may slightly affect the apprehension of salient objects. For category 4+, there is $78\%$ agreement, and about $13\%$ of images in this category are classified as category 0.

In Fig.~\ref{fig:sbt_amt_diff}, we show sample images that are consistently labeled by all three subjects in our offline subitizing test as a different category than labeled by AMT workers. We find some labeling discrepancy may be attributed to the fact that objects at the image center tend to be thought of as more salient than other ones given a short viewing time (see images in the top row of Fig.~\ref{fig:sbt_amt_diff}). In addition, some images with many foreground objects (far above the subitizing limit of 4 ) are labeled as 4+ by AMT workers, but they tend to be labeled as category 0 in our offline subitizing test (see the middle and right images at the bottom row in Fig.~\ref{fig:sbt_amt_diff}).

 Despite the labeling discrepancy on a small proportion of the sampled images, limiting the viewing time to a fraction of a second does not significantly decrease the inter-subject consistency or change the answers to the number of salient objects on most test images. We thereby believe the proposed SOS dataset is valid. The per-class accuracy shown in Fig.~\ref{fig:human_conf} (percentage numbers in diagonal cells) can be interpreted as an estimate of the human performance baseline on our dataset.

\section{Salient Object Subitizing by Convolutional Neural Network}

Subitizing is believed to be a holistic sense of the number of objects in a visual scene. This visual sense can discriminate between the visual patterns possessed by different numbers of objects in an image \citep{jansen2014role,mandler1982subitizing,clements1999subitizing,boysen2014development}. This inspires us to propose a learning-based discriminative approach to address the SOS problem, without resorting to any object localization or counting process. In other words, we aim to train image classifiers to predict the number of salient objects in a image.

%To this end, we will first benchmark several traditional image feature representations as well as the state-of-the-art CNN-based approach. This will provide the first set of baselines for the SOS problem. Then we will investigate how to leverage synthetic images to improve the training of the CNN model.
%
%\subsection{Feature Representation}
%
%Although salient objects can have dramatically different appearance in color, texture and shape, we expect that global geometric information can be used to differentiate images with different numbers of salient objects. Therefore, we evaluate HOG \citep{dalal2005histograms}, GIST \citep{torralba2003context} and Improved Fisher Vectors (IFV) \citep {perronnin2010improving} of dense SIFT \citep{lowe2004distinctive}, all of which are gradient-based features and have been used for image classification. We also evaluate a spatial pyramid feature of saliency maps, in the hope that saliency maps may provide information about the composition of the foreground.
%
Encouraged by the remarkable progress made by the CNN models in computer vision \citep{girshick14CVPR,krizhevsky2012imagenet,razavian2014cnn,sermanet2013overfeat}, we use the CNN-based method for our problem. \cite{girshick14CVPR} suggest that given limited annotated data, fine-tuning a pre-trained CNN model can be an effective and highly practical approach for many problems. Thus, we adopt fine-tuning to train the CNN SOS model.
%
%We now provide more detailed descriptions about the feature representations we are going to benchmark.
%

We use the GoogleNet architecture \citep{szegedy2015going}, which has significantly fewer parameters than the AlexNet model in our previous SOS paper \citep{zhang2015salient}. However, GoogleNet is shown to substantially outperform AlexNet in image classification tasks and it also compares favorably with the widely used the VGG16 \citep{simonyan2014very} architecture in terms of speed and classification accuracy.
We fine-tune the GoogleNet model pre-trained on ImageNet \citep{ILSVRCarxiv14} using Caffe \citep{jia2014caffe}. The output layer of the pre-trained GoogleNet model is replaced by a fully connected layer which outputs a 5-D score vector for the five categories: 0, 1, 2, 3 and 4+. We use the Softmax loss and the SGD solver of Caffe to fine-tune all the parameters in the model. More training details are provided in Sec.~\ref{sec:exp}.

%using Caffe  \citep{jia2014caffe}
% We use Caffe  \cite{jia2014caffe} for fine-tuning the CNN model pre-trained on ImageNet \cite{ILSVRCarxiv14}. Images are resized to $256\times 256$ regardless of of their original aspect ratios. The top-left, top-right, bottom-left and bottom-right $227\times 227$ crops of a image are used to augment the training data. We use Caffe's default setting for training the CNN model of \cite{krizhevsky2012imagenet}, but reduce the starting learning rate to 0.001 as in \cite{girshick14CVPR}. We stop tuning after around 30 epochs, as the training loss no longer decreases.

\subsection{Leveraging Synthetic Images for CNN Training}

Collecting and annotating real image data is a rather expensive process. Moreover, the collected data may not have a balanced distribution over all the categories. In our SOS dataset, over 2/3 images belong to the ``0" or ``1" category. For categories with insufficient data, the CNN model training may suffer from overfitting and lead to degraded generalizability of the CNN model.

Leveraging synthetic data can be a economical way to alleviate the burden of image collection and annotation \citep{stark2010back,sun2014virtual,jaderberg2014synthetic}. In particular, some recent works \citep{jaderberg2014synthetic,peng2015learning} successfully exploit synthetic images to train modern CNN models for image recognition tasks. While previous works focus on generating realistic synthetic images (\emph{e.g.}\ using 3D rendering techniques \citep{peng2015learning}) to train CNN models with zero or few real images data, our goal is to use synthetic images as an auxiliary source to improve the generalizability of the learned CNN model.

We adopt a convenient \emph{cut-and-past} approach to generate synthetic SOS image data. Given a number $N$ in the range of 1-4, a synthetic image is generated by pasting $N$ cutout objects on a background scene image. Cutout objects can be easily obtained from existing image datasets with segmentation annotations or image sources with isolated object photos (\emph{e.g.}\ stock image databases). In this work, we use the public available salient object dataset THUS10000 \citep{ChengPAMI} for generating cutout objects and the SUN dataset \citep{xiao2010sun} as the source for background images. The THUS10000 dataset covers a wide range of object categories so that we can obtain sufficient variations in the shape and appearance of foreground objects.

In THUS10000, an image may contain multiple salient objects and some of them are covered by a single segmentation mask. To generate consistent synthetic SOS image data, we automatically filter out this type of images using the CNN SOS model trained on real data. To do this, we remove the images whose confidence scores for containing one salient object are less than 0.95. Similarly, we filter out the images with salient objects from the SUN dataset, using a score threshold of 0.95 for containing no salient object.

When generating a synthetic image, we randomly choose a background image and resize it to 256$\times$256 regardless of its original aspect ratio. Then, we pick a cutout object and generate a reference object by resizing it to a randomly generated scale relative to 256 based on the largest dimension of the object. The reference scale is uniformly sampled in the range $[0.4,0.8]$. After that, we apply random horizontal flipping and mild geometric transforms (scaling and rotation) on the reference object each time we past a copy of it to a random position on the background image. Mild scalings are uniformly sampled in the range $[0.85,1.15]$ and mild rotations are uniformly sampled in the angular range $[-10,10]$ degrees. The synthetic image contains $N$ ($N\in [1,4]$) copies of the same cutout object. Pasting different cutout objects together is empirically found inferior to our method, probably because some cutout objects may appear more salient than the other ones when they are put together, resulting in images that visually inconsistent with the given number. Finally, we reject this image if any of the pasted objects is occluded by more 50\% of its area.

%\footnote{We have also tried pasting different cutout objects together, but that often leads to synthetic images that are visually inconsistent with the given number, because some cutout objects may appear more salient than the others when they are put together. Empirical results also show that using this type synthetic image does not improve the CNN training.}

\begin{figure}
\centering
	\includegraphics[width = 1\linewidth]{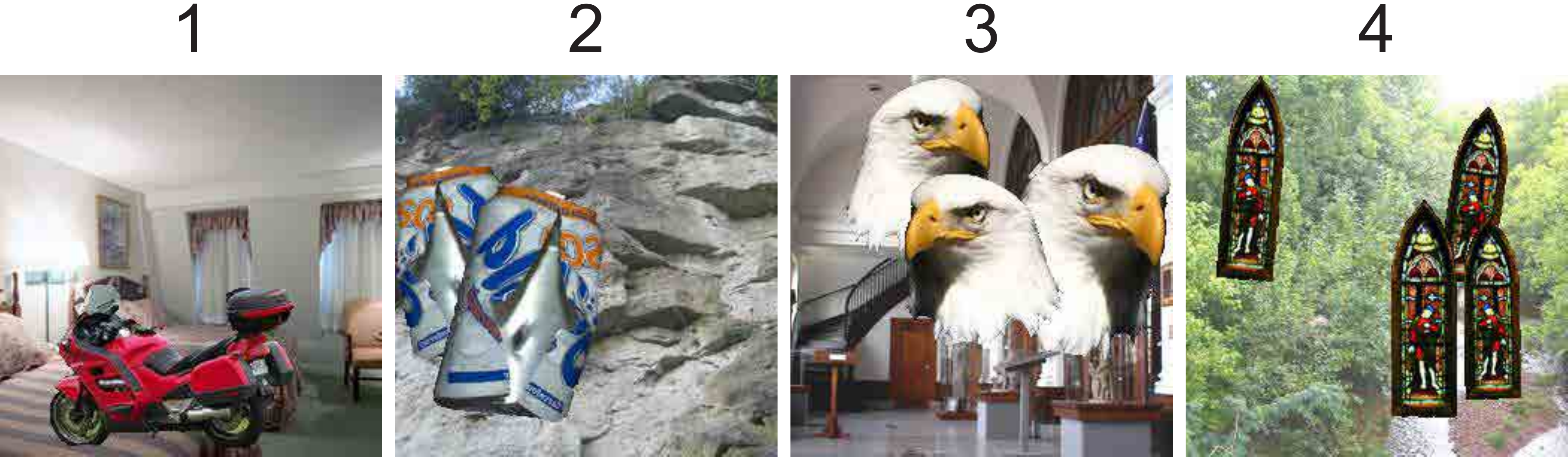}
    \caption{Sample synthetic images with the given numbers of salient objects on the top. Although the synthetic images look rather unrealistic, they are quite visually consistent with the given numbers of salient objects. By pre-training the CNN SOS model on these synthetic images, we expect that the CNN model can better learn the intra-class variations in object category, background scene type, object position and inter-object occlusion.}\label{fig:syn}
\end{figure}

Example synthetic images are shown in Fig.~\ref{fig:syn}. Our synthetic images look rather unrealistic, since we do not consider any contextual constraints between scene types and object categories. However, for the SOS task, these images often look quite consistent with the given numbers of salient objects. We expect that our CNN model should learn generic features for SOS irrespective of semantics of the visual scenes. Thus, these synthetic images may provide useful intra-class variations in object category, background scene type, as well as object position and inter-object occlusion.

To leverage the synthetic images, we fine-tune the CNN model on the synthetic data before fine-tuning on the real data.
The two-stage fine-tuning scheme can be regarded as a domain adaptation process, which transfers the learned features from the synthetic data domain to the real data domain. Compared with combining the real and synthetic images into one training set, we find that our two-stage fine-tuning scheme works significantly better (see Sec.~\ref{sec:exp}).

\section{Experiments}\label{sec:exp}

\begin{table*}
\caption{Average Precision (\%) of compared methods. The best scores are shown in bold. The training and the testing are repeated for five times for all CNN-based methods, and mean and std of the AP scores are reported.}
\label{tab:ap}       % Give a unique label
\centering
\begin{tabular}{rC{1.3cm}C{1.3cm}C{1.3cm}C{1.3cm}C{1.3cm}C{1.3cm}}
\toprule
            & 0 & 1 & 2 & 3 & 4+ & mean  \\\cmidrule{1-7}
Chance      & 27.5 & 46.5 & 18.6 & 11.7 & 9.7  & 22.8\\\cmidrule{1-1}
%SalCount    &  &    & & & & \\
SalPyr      & 46.1 & 65.4 & 32.6 & 15.0 & 10.7 & 34.0 \\
HOG         & 68.5 & 62.2 & 34.0 & 22.8 & 19.7 & 41.4 \\
GIST        & 67.4 & 65.0 & 32.3 & 17.5 & 24.7 & 41.4 \\
SIFT+IVF    & 83.0 & 68.1 & 35.1 & 26.6 & 38.1 & 50.1 \\\cmidrule{1-1}
CNN\_woFT    & 92.2$\pm$0.2 & 84.4$\pm$0.2 & 40.8$\pm$1.9 & 34.1$\pm$2.7 & 55.2$\pm$0.6 & 61.3$\pm$0.2 \\
CNN\_FT         & \textbf{93.6}$\pm$0.3 & \textbf{93.8}$\pm$0.1 & 75.2$\pm$0.2 & 58.6$\pm$0.8 & 71.6$\pm$0.5 & 78.6$\pm$0.2 \\\cmidrule{1-1}
CNN\_Syn     & 79.2$\pm$0.5 & 85.6$\pm$0.2 & 37.4$\pm$0.8 & 34.8$\pm$2.6 & 33.0$\pm$1.1 & 54.0$\pm$0.6 \\
CNN\_Syn\_Aug     & 92.1$\pm$0.4 & 92.9$\pm$0.1 & 75.0$\pm$0.4 & 58.9$\pm$0.6 & 69.8$\pm$0.8 & 77.8$\pm$0.3 \\
CNN\_Syn\_FT     & \textbf{93.5}$\pm$0.1 & \textbf{93.8}$\pm$0.2 & \textbf{77.4}$\pm$0.3 & \textbf{64.3}$\pm$0.2 & \textbf{73.0}$\pm$0.5 & \textbf{80.4}$\pm$0.2 \\
\bottomrule
\end{tabular}
\end{table*}

\subsection{Experimental Setting}

For training and testing, we randomly split the SOS dataset into a training set of 10,966 images (80\% of the SOS dataset) and a testing set of 2741 images.

%We train linear SVM classifiers for GIST, HOG, IFV and the saliency map pyramid feature (SalPyr). The hyper-parameters of the SVM are determined via 5-fold validation.

\textbf{CNN model training details.} For fine-tuning the GoogleNet CNN model, images are resized to $256\times 256$ regardless of their original aspect ratios. Standard data augmentation methods like horizontal flipping and cropping are used. We set the batch size to 32 and fine-tune the model for 8000 iterations. The fine-tuning starts with a learning rate of 0.001 and  we multiply it by 0.1 every 2000 iterations. At test time, images are resized to $224\times 224$ and the output softmax scores are used for evaluation.

For pre-training using the synthetic images, we generate 5000 synthetic images for each number in 1-4. Further increasing the number of synthetic images does not increase the performance. We also include the real background images (category ``0") in the pre-training stage. The same model training setting is used as described above. When fine-tuning using the real data, we do not reset the parameters of the top fully-connected layer, because we empirically find that it otherwise leads to slightly worse performance.

\textbf{Compared methods.}
We evaluate our method and several baselines as follows.
\begin{itemize}
  \item {CNN\_Syn\_FT}: The full model fine-tuned using the two-stage fine-tuning scheme with the real and synthetic image data.
  \item {CNN\_Syn\_Aug}: The model fine-tuned on the union of the synthetic and the real data. This baseline corresponds to the data augmentation scheme in contrast to the two-stage fine-tuning scheme for leveraging the synthetic image data. This baseline is to validate our two-stage fine-tuning scheme.
  \item {CNN\_FT}: The CNN model fine-tuned on the real image data only.
  \item {CNN\_Syn}: The CNN model fine-tuned on the synthetic images only. This baseline reflects how close the synthetic images are to the real data.
  \item {CNN\_wo\_FT}: The features of the pre-trained GoogleNet without fine-tuning. For this baseline, we fix the parameters of all the hidden layers during fine-tuning. In other words, only the output layer is fine-tuned.
\end{itemize}

Furthermore, we benchmark several commonly used image feature representations for baseline comparison. For each feature representation, we train a one-vs-all multi-class linear SVM classifier on the training set. The hyper-parameters of the SVM are determined via five-fold cross-validation.

\begin{itemize}
  \item {GIST.} The GIST descriptor \citep{torralba2003context} is computed based on 32 Gabor-like filters with varying scales and orientations. We use the implementation by \cite{torralba2003context} to extract a 512-D GIST feature, which is a concatenation of averaged filter responses over a $4\times 4$ grid.
  \item {HOG.} We use the implementation by  \cite{felzenszwalb2010object} to compute HOG features. Images are first resized to $128\times 128$, and HOG descriptors are computed on a $16\times 16$ grid, with the cell size being $8\times 8$. The HOG features of image cells are concatenated into a 7936-D feature. We have also tried combining HOG features computed on multi-scale versions of the input image, but this gives little improvement.
  \item {SIFT with the Improved Fisher Vector Encoding (SIFT+IVF).} We use the implementation by \cite{Chatfield11}. The codebook size is 256, and the dimensionality of SIFT descriptors is reduced to 80 by PCA. Hellinger's kernel and L2-normalization are applied for the encoding. Weak geometry information is captured by spatial binning using $1\times 1$, $3 \times 1$ and $2 \times 2$ grids.
      %Readers are referred to \cite{Chatfield11} for more details.
      To extract dense SIFT, we use the VLFeat \cite{vedaldi08vlfeat} implementation. Images are resized to $256\times 256$, and a $8\times 8$ grid is used to compute a 8192-D dense SIFT feature, with a step size of 32 pixels and a bin size of 8 pixels. Similar to HOG, combining SIFT features of different scales does not improve the performance.
  \item {Saliency map pyramid (SalPyr).}  We use a state-of-the-art CNN-based salient object detection model \citep{zhao2015saliency} to compute a saliency map for an image. Given a saliency map, we construct a spatial pyramid of a $8\times 8$ layer and a $16\times 16$ layer. Each grid cell represents the average saliency value within it. The cells of the spatial pyramid are then concatenated into a 320-D vector.
\end{itemize}

\textbf{Evaluation metric.}
We use average precision (AP) as the evaluation metric. We use the implementation provided in the VOC07 challenge \cite{pascal-voc-2007} to calculate AP. For each the CNN-based method, we repeat the training for five times and report both the mean and the standard deviation (std) of the AP scores. This will give a sense of statistical significance when interpreting the difference between CNN baselines.

\subsection{Results}
The AP scores of different features and CNN baselines are reported in Table~\ref{tab:ap}. The baseline Chance in Table~\ref{tab:ap} refers to the performance of random guess. To evaluate the random guess baseline, we generate random confidence scores for each category, and report the average AP scores over 100 random trials.

All methods perform significantly better than random guess in all categories. Among manually crafted features, SalPyr gives the worst mean AP (mAP) score, while SIFT+IFV performs the best, outperforming SalPyr by 16 absolute percentage points in mAP. SIFT+IFV is especially more accurate than other non-CNN features in identifying images with 0 and 4+ salient objects.

The CNN feature without fine-tuning (CNN\_wo\_FT) outperforms SIFT+IFV by over 10 absolute percentage points in mAP. Fine-tuning (CNN\_FT) further improves the mAP score by 17 absolute percentage points, leading to a mAP score of 78.6\%. CNN\_wo\_FT attains comparable performance to CNN\_FT in identifying background images, while it is significantly worse than CNN\_FT in the other categories. This suggests that the CNN feature trained on ImageNet is good for inferring the presence of salient objects, but not very effective at discriminating images with different numbers of salient objects.

\begin{figure}
\centering
	\includegraphics[width = 1\linewidth]{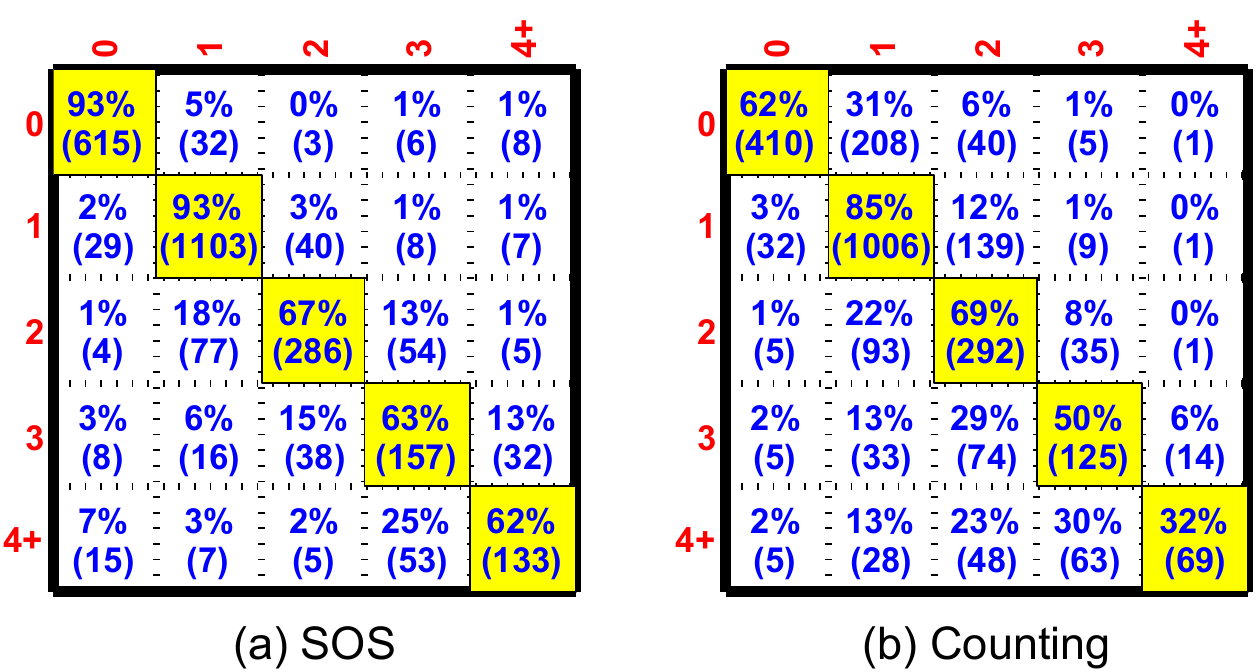}
    \caption{Subitizing \emph{vs.}\ counting. (a) Confusion matrix of our CNN SOS method CNN\_Syn\_FT. Each row corresponds to a groundtruth category. The percentage reported in each cell is the proportion of images of the category A (row number) labeled as category B (column number). (b) Confusion matrix of counting using the salient object detection method by \cite{zhang2015SOD}.}\label{fig:confmat}
\end{figure}

\begin{figure*}
\centering
	\includegraphics[width = 1\linewidth]{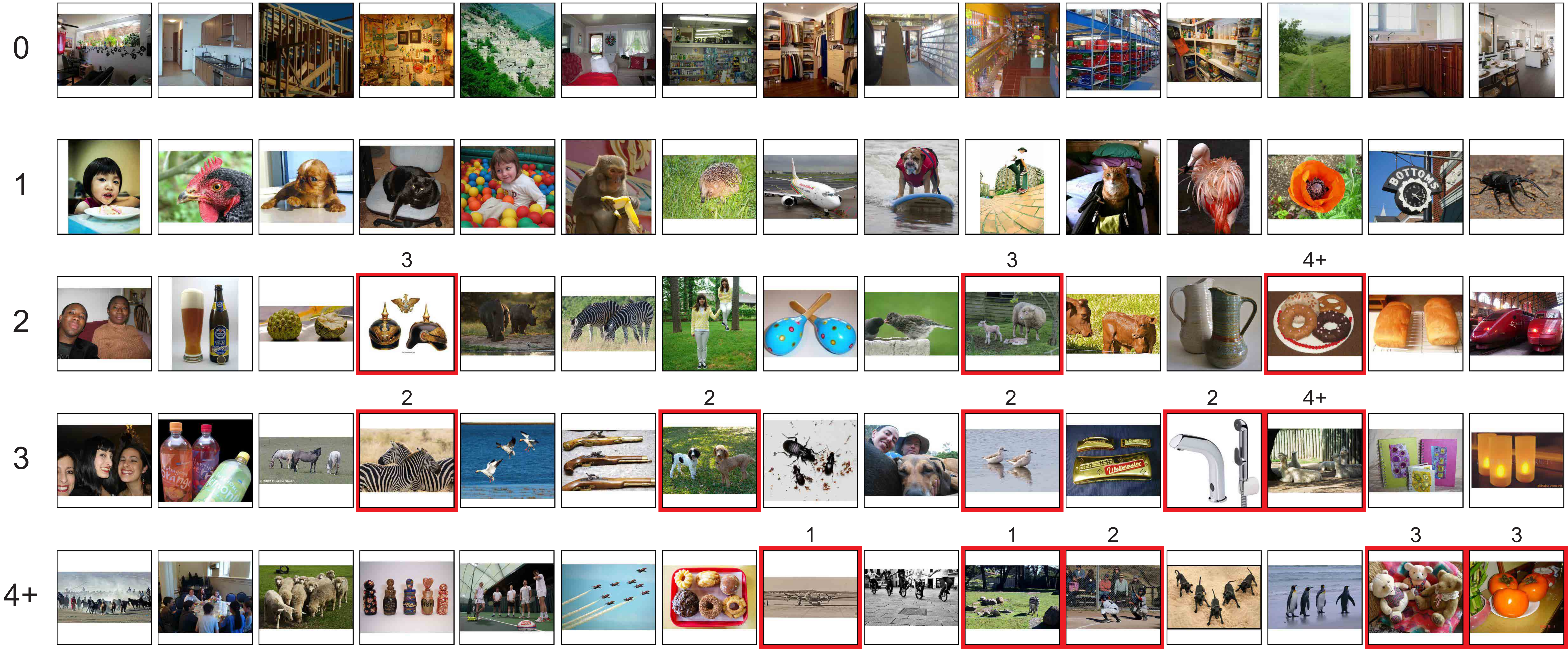}
    \caption{Sample results among the top 100 predictions for each category by our CNN SOS method CNN\_Syn\_FT. The images are listed in descending order of confidence. False alarms are shown with red borders and groundtruth labels at the top.}\label{fig:demo}
\end{figure*}

Pre-fine-tuning using the synthetic images (CNN\_Syn\_FT) further boosts the performance of CNN\_FT by about 2 absolute percentage points in mAP. The performance is improved in category ``2", ``3" and ``4+", where training images are substantially fewer than categories ``0" and ``1". In particular, for category ``3" the AP score is increased by about 6 absolute percentage points. The usefulness of the synthetic images may be attributed to the fact they can provide more intra-class variations in object category, scene type and the spatial relationship between objects. This is especially helpful when there is not enough real training data to cover the variations.

Using synthetic images alone (CNN\_Syn) gives reasonable performance, a mAP score of 54.0\%. It outperforms SIFT+IVF, the best non-CNN baseline trained on the real data. However, it is still much worse than the CNN model trained on the real data. This gives a sense of the domain shift between the real and the synthetic data. Directly augmenting the training data with the synthetic images does not improve and even slightly worsens the performance (compare CNN\_Syn\_Aug and CNN\_FT in Table~\ref{tab:ap}). We believe that this is due to the domain shift and our two-stage fine-tuning scheme can better deal with this issue.

Fig.~\ref{fig:confmat} (a) shows the confusion matrix for our best method CNN\_Syn\_FT. The percentage reported in each cell represents the proportion of images of category A (row number) classified as category B (column number).
The accuracy (recall) of category ``0" and ``1" is both about 93\%, which is close to the human accuracy for these categories in our human subitizing test (see Fig.~\ref{fig:human_conf}). For the remaining categories, there is still a considerable gap between human and machine performance. According to Fig.~\ref{fig:confmat} (a), our SOS model tends to make mistaks by misclassifying an image into a nearby category. Sample results are displayed in Fig.~\ref{fig:demo}. Despite the diverse object appearance and image background, our SOS model gives reasonable performance.

\subsection{Analysis}

To gain a better understanding of our SOS method, we further investigate the following questions.

\textbf{How does subitizing compare to counting?} Counting is a straightforward way of getting the number of items. To compare our SOS method with a counting-by-detection baseline, we use a state-of-the-art salient object detection method designed for unconstrained images \citep{zhang2015SOD}. This unconstrained salient object detection method, denoted as USOD, leverages a CNN-based model for bounding proposal generation, followed by a subset optimization method to extract a highly reduced set of detection windows. A parameter of USOD is provided to control the operating point for the precision-recall tradeoff. We pick an operating point that gives the best F-score\footnote{The F-score is computed as $\frac{2RP}{(R+P)}$, where $R$ and $P$ denote recall and precision respectively.} on the Multi-Salient-Object (MSO) dataset \citep{zhang2015salient} in this experiment.

The confusion matrix of the counting baseline is shown in Fig.~\ref{fig:confmat} (b). Compared with the SOS method (see Fig.~\ref{fig:confmat} (a)), the counting baseline performs significantly worse in all categories except ``2". In particular, for ``0" and ``4+", the counting baseline is worse than the SOS method by about 30 absolute percentage points. This indicates that for the purpose of number prediction, the counting-by-detection approach can be a suboptimal option. We conclude that there are at least two reasons for this outcome. First, it is difficult to pick a fixed score threshold (or other equivalent parameters) of an object detection system that works best for all images. Even when an object detector gives a perfect ranking of window proposals for each image, the scores may not be well calibrated across different images. Second, the post-processing step for extracting detection results (\emph{e.g.}\ non-maximum suppression) is based on the idea of suppressing severely overlapping windows. However, this spatial prior about detection windows can be problematic when significant inter-object occlusion occurs. In contrast, our SOS method bypass the detection process and discriminates between different numbers of salient objects based on holistic cues.

\begin{table}
\centering
\caption{Mean average precision (\%) scores for different CNN architectures. Training and test are run for five times and the mean and the std of mAP scores are reported.}
\label{tab:arch}
\begin{tabular}{rccc}
\toprule
                  & AlexNet & VGG16 & GoogleNet \\\cmidrule{1-4}
w/o Syn. Data     & 70.1$\pm$0.2 & 77.5$\pm$0.3 & 78.6$\pm$0.2 \\
with Syn. Data    & 71.6$\pm$0.5 & 80.2$\pm$0.3 & 80.4$\pm$0.3\\\bottomrule
\end{tabular}
\end{table}

\textbf{How does the CNN model architecture affect the performance?} Besides GoogleNet, we evaluate another two popular architectures, AlexNet \citep{krizhevsky2012imagenet} and VGG16 \citep{simonyan2014very}. The mAP scores with and without using synthetic images are summarized in Table~\ref{tab:arch} for each architecture. VGG16 and GoogleNet have very similar performance, while AlexNet performs significantly worse. Pre-training using synthetic images has a positive effect on all these architectures, indicating that it is generally beneficial to leverage synthetic images for this task. The baseline of AlexNet without synthetic image can be regarded as the best model reported by \cite{zhang2015salient}. In this sense, our current best method using GoogleNet and synthetic image outperforms the previous best model by 10 absolute percentage points. Note that the training and testing image sets used by \cite{zhang2015salient} are subsets of the training and testing sets of our expanded SOS dataset. Therefore, the scores reported by \cite{zhang2015salient} are not comparable to the scores in this paper\footnote{When evaluated on the test set used by \cite{zhang2015salient}, our best method GoogleNet\_Syn\_FT achieves a mAP score of 85.0\%}.

\begin{table}
\caption{The effect of using the synthetic images when different numbers of real data are used in CNN training. For each row, the same set of synthetic images are used. Training and test are run for five times and the mean and the std of mAP scores are reported. By using the synthetic images, competitive  performance is attained even when the size of the real data is significantly reduced.}
\label{tab:train}
\centering
\begin{tabular}{rcc}
\toprule
                   & w/o syn.         & with syn. \\\cmidrule{1-3}
   25\% real data  & 71.6$\pm$0.2     &  76.3$\pm$0.4\\
   50\% real data  & 75.3$\pm$0.3     &  78.2$\pm$0.4\\
   100\% real data & 78.6$\pm$0.2     & 80.4$\pm$0.3 \\\bottomrule
\end{tabular}
\end{table}

\textbf{Does the usage of synthetic images reduce the need for real data?} To answer this question, we vary the amount of real data used in the training, and report the mAP scores in Table \ref{tab:train}. We randomly sample 25\% and 50\% of the real data for training the model. This process is repeated for five times. When fewer real data are used, the performance of our CNN SOS method declines much slower with the help of the synthetic images. For example, when only 25\% real data are used, leveraging the synthetic images can provide an absolute performance gain of about 5\% in mAP, leading to a mAP score of 76\%. However, without using the synthetic images, doubling the size of the training data (50\% real data) only achieves a mAP score of 75\%. This suggests that we can achieve competitive performance at a much lower cost at data collection by leveraging the synthetic images.

\begin{figure*}
\centering
	\includegraphics[width = 1\linewidth]{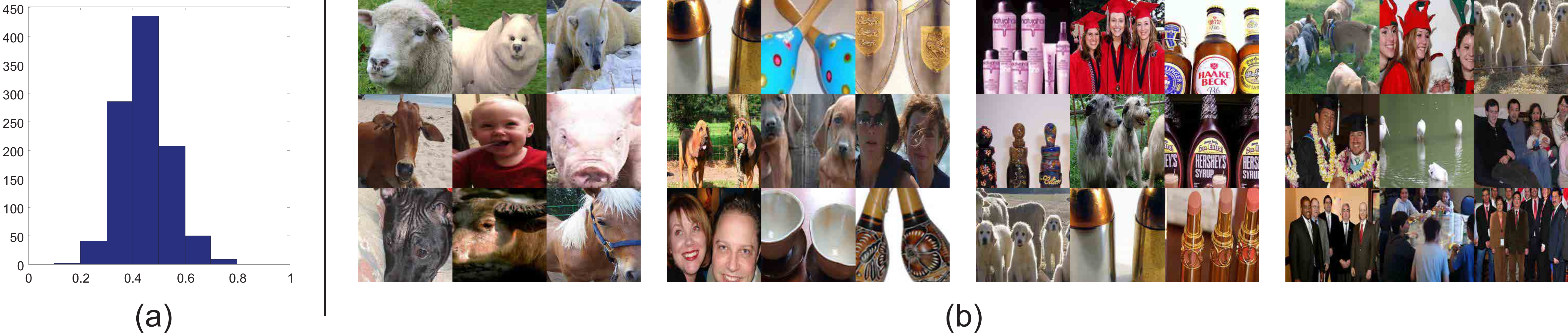}
    \caption{Feature visualization of the inception\_5b/output layer in our GoogleNet\_Syn\_FT model. We aim to visualize the new feature representations learned from our SOS data. (a) shows the histogram of $S_i$, which measures how distinct a feature channel of our model is from the feature representations of the original ImageNet model (see text for more details). Lower values of $S_i$ indicates higher distinctness, and we choose those feature channels with $S_i < 0.3$ for visualization (b) shows the visualization of some new feature representations learned by our SOS model. Each block displays the top nine image patches in our SOS test set that correspond to the highest feature activations for a novel feature channel. These visualization results suggest that our CNN model has learned some category-independent and discriminative features for SOS. For example, the first block corresponds to a feature about a close-up face, and the second block shows a feature of a pair of objects appearing side by side. }\label{fig:vis}
\end{figure*}

\textbf{What is learned by the CNN model?} By fine-tuning the pre-trained CNN model, we expect that the CNN model will learn discriminative and generalizable feature representations for subitizing. To visualize the new feature representations learned from our SOS data, we first look for features that are substantially distinct from the ones of the original network trained on ImageNet. For GoogleNet, we consider the output layer of the last inception unit (inception\_5b/output), which has 1024 feature channels. For each feature channel of this layer, we use the maximum activation value on an image to rank the images in the SOS test set. We hypothesize that if two feature channels represent similar features, then they should result in similar image rankings. Given the $i$-th feature channel of this layer in GoogleNet\_Syn\_FT, we compute the maximum Spearman's rank correlation coefficient between its image ranking $R_i$ and the image ranking $\widehat{R}_j$ using the $j$-th channel of the original GoogleNet:
\begin{equation}
    S_i = \max_{j=1,2\cdots,1024} \rho(R_i, \widehat{R}_j),
\end{equation}
where $\rho$ denotes Spearman's rank correlation coefficient, whose range is $[-1,1]$. A low value of $S_i$ means that the $i$-th feature channel of our fine-tuned model gives a very different image ranking than any feature channels from the original CNN model. In our case, none of the values of $S_i$ is negative. Fig.~\ref{fig:vis} (a) shows the histogram of $S_i$. We choose the feature channels with $S_i$ values less than $0.3$ as the most novel features learned from the SOS data.

After that, we visualize each of the novel feature channels by showing the top nine image patches in our SOS test set that correspond to the highest feature activations for that channel. The spatial resolution of inception\_5b/output is 7$\times$7. For an activation unit on the 7$\times$7 map, we display the image patch corresponding to the receptive field of the unit. Since the theoretic receptive field of the unit is too large, we restrict the image patch to be 60\% of the size (0.6W$\times$0.6H) of the whole image.

Fig.~\ref{fig:vis} (b) shows the visualization results of some of the novel feature representations learned by our CNN SOS model. We find that these newly learned feature representations are not very sensitive to the categories of the objects, but they capture some general visual patterns related to the subitizing task. For example, in Fig.~\ref{fig:vis} (b), the feature corresponding to the first block is about a close-up face of either a person or an animal. Detecting a big face at this scale indicates that the image is likely to contain only a single dominant object. The feature corresponding to the second block is about a pair of objects appearing side by side, which is also a discriminative visual pattern for identifying images with two dominant objects. These visualization results suggest that our CNN model has learned some category-independent and discriminative features for SOS.

\begin{table*}
\caption{Cross-category generalisation test. The CNN-LOO refers to the AP scores (\%) on the unseen object category. CNN-Full serves as an upper bound of the performance when the images of that object category are used in the training (see text for more details). The number following each category name is the number of images with that category label.}
\label{tab:CV}       % Give a unique label
\centering
\begin{tabular}{rcC{1.1cm}C{1.1cm}C{1.1cm}C{1.1cm}C{1.1cm}C{1.1cm}}
\toprule
      &     & 0 & 1 & 2 & 3 & 4+ & mean  \\\cmidrule{2-8}
  \multirow{ 3}{*}{animal (4101)} &   Chance      & 16.6 & 53.6 & 21.1 & 12.6 & 8.8  & 22.5\\
  &   CNN-LOO      & 89.3$\pm$0.2 & 87.2$\pm$0.3 & 42.8$\pm$1.0 & 36.9$\pm$2.6 & 58.3$\pm$1.0  & 62.9$\pm$0.5\\
  &   CNN-Full      & 95.0$\pm$1.7 & 94.8$\pm$0.4 & 72.8$\pm$2.0 & 57.9$\pm$2.8 & 71.8$\pm$4.0  & 78.5$\pm$1.3\\\cmidrule{2-8}
  \multirow{ 3}{*}{food (372)} &   Chance      & 67.6 & 16.9 & 8.1 & 13.1 & 8.2  & 22.8\\
  &   CNN-LOO      & 95.7$\pm$0.2 & 70.8$\pm$1.3 & 50.3$\pm$0.8 & 56.8$\pm$1.3 & 39.7$\pm$1.4  & 62.7$\pm$0.5\\
  &   CNN-Full      & 97.7$\pm$0.4 & 85.9$\pm$7.2 & 61.1$\pm$11.2 & 67.8$\pm$12.4 & 62.8$\pm$8.3  & 75.1$\pm$4.1\\\cmidrule{2-8}
  \multirow{ 3}{*}{people (3786)} &   Chance      & 17.5 & 50.7 & 21.7 & 10.9 & 13.1  & 22.8\\
  &   CNN-LOO      & 86.7$\pm$0.3 & 84.9$\pm$0.5 & 47.6$\pm$0.5 & 31.6$\pm$1.3 & 56.7$\pm$1.2  & 61.5$\pm$0.5\\
  &   CNN-Full      & 94.4$\pm$1.3 & 94.8$\pm$0.7 & 82.5$\pm$1.0 & 62.8$\pm$6.1 & 83.9$\pm$2.8  & 83.7$\pm$1.3\\\cmidrule{2-8}
  \multirow{ 3}{*}{vehicle (1150)} &   Chance      & 40.6 & 56.1 & 8.3 & 3.4 & 4.4  & 22.6\\
  &   CNN-LOO      & 91.0$\pm$0.3 & 92.2$\pm$0.3 & 42.4$\pm$2.2 & 16.3$\pm$0.9 & 47.4$\pm$0.9  & 57.9$\pm$0.4\\
  &   CNN-Full      & 96.1$\pm$0.7 & 96.1$\pm$0.7 & 62.2$\pm$9.2 & 25.6$\pm$14.2 & 55.4$\pm$20.6  & 67.1$\pm$6.4\\\cmidrule{2-8}
  \multirow{ 3}{*}{other (1401)} &   Chance      & 36.4 & 35.4 & 14.8 & 18.6 & 11.2  & 23.3\\
  &   CNN-LOO      & 87.0$\pm$0.4 & 78.0$\pm$0.7 & 56.7$\pm$0.4 & 49.9$\pm$0.9 & 50.2$\pm$0.8  & 64.4$\pm$0.4\\
  &   CNN-Full      & 93.4$\pm$0.4 & 90.5$\pm$2.5 & 70.8$\pm$7.2 & 63.0$\pm$3.2 & 60.2$\pm$8.3  & 75.6$\pm$2.8\\
\bottomrule
\end{tabular}
\end{table*}

\textbf{How does the SOS method generalize to unseen object categories?} We would like to further investigate how our CNN SOS model can generalize to unseen object categories. To get category information for the SOS dataset, we ask AMT workers to label the categories of dominant objects for each image in our SOS dataset. We consider five categories: ``animal", ``food", ``people", ``vehicle" and ``other". An image may contain multiple labels (\emph{e.g.}\ an image with an animal and a person). For each image, we collect labels from three different workers and use the majority rule to decide the final labels.

To test the generalizability of our CNN model to unseen object categories, we use the Leave-One-Out (LOO) approach described as follows. Given category $\mathcal{A}$, we remove all the images with the label $\mathcal{A}$ from the original training set, and use them as the testing images. The original test images for ``0" are also included. Two other baselines are provided. The first is a chance baseline, which refers to the performance of random guess. We generate random confidence scores for each category, and report the average AP scores over 100 random trials. Note that we have class imbalance in the test images, so the AP scores of random guess tend to be higher for categories with more images. The second baseline reflects the performance for category $\mathcal{A}$ when full supervision is available. We use five-fold cross-validation to evaluate this baseline. In each fold, $1/5$ of the images with the label $\mathcal{A}$ are used for testing, and all the remaining images are used for training. The average AP scores are reported.
%We randomly divide the test set for category $\mathcal{A}$ into five equal subsets, and use one of the subsets as the testing images. Images with the label $\mathcal{A}$ in the rest of the subsets are added to the training set. The average AP scores of this five-fold cross-validation are reported.
In this experiment, we do not use the synthetic images because they do not have category labels.

The results are reported in Table~\ref{tab:CV}. For each category, the CNN model trained without that category (CNN-LOO) gives significantly better performance than the Chance baseline. This validates that the CNN model can learn category-independent features for SOS and it can generalize to unseen object categories to some extent. Training with full supervision (CNN-Full) further improves over CNN-LOO by a substantial margin, which indicates that it is still important to use a training set that covers a diverse set of object categories.

\section{Applications}

\subsection{Salient Object Detection}

In this section, we demonstrate the usefulness of SOS for unconstrained salient object detection \citep{zhang2015SOD}. Unconstrained salient object detection aims to detect salient objects in unconstrained images where there can be multiple salient objects or no salient objects. Compared with the constrained setting, where there exists one and only one salient object, the unconstrained setting pose new challenges of handling background images and determining the number of salient objects. Therefore, SOS can be used to cue a salient object detection method to suppress the detection or output the right number of detection windows for unconstrained images.

Given a salient object detection method, we leverage our CNN SOS model by a straightforward approach. We assume that the salient object detection method provides a parameter (\emph{e.g.} the threshold for the confidence score) for trade-off between precision and recall. We call this parameter as a PR parameter. For an image, we first predict the number of salient objects $N$ using our CNN SOS model, then we use grid search to find such a value of the PR parameter that no more than $N$ detection windows are output.

\textbf{Dataset.} Most existing salient object detection datasets lack background images or images containing multiple salient objects. In this experiment, We use the Multi-Salient-Object (MSO) dataset \citep{zhang2015salient}. The MSO dataset has 1224 images, all of which are from the test set of the SOS dataset, and it has a substantial proportion of images that contain no salient object or multiple salient objects.

\textbf{Compared methods.} We test our SOS model on the unconstrained object detection method proposed (denoted as USOD) by \cite{zhang2015SOD}, which achieves state-of-the-art performance on the MSO dataset. The baseline USOD method is composed of a CNN-based object proposal model and a subset optimization formulation for post-processing the bounding box proposals. We use an implementation provided by \cite{zhang2015SOD}, which uses the GoogleNet architecture for proposal generation. The USOD method provides a PR parameter to control the number of detection windows. We use the predicted number by our SOS model to cue USOD, and denote this method as USOD+SOS. We also use the groundtruth number to show the upper-bound of the performance gain using subitizng, and denote this baseline as USOD+GT.
%In addition, we test an variant of USOD by replacing the proposal model of USOD with the Multi-Box object proposal model \citep{erhan2014scalable}. We denote this baseline as MBox. The proposal confidence scores generated by the Mutli-Box model is less calibrated than the original proposal model of USOD, since the Mutli-Box model is trained for

\textbf{Evaluation metrics.} We report the precision, the recall and the F-measure. The F-measure is calculated as $2\frac{PR}{P+R}$, where $P$ and $R$ denote the precision and the recall respectively. For the baseline USOD method, we tune its PR parameter so that the its F-measure is maximized.

\begin{figure}
\centering
	\includegraphics[width = 0.65\linewidth]{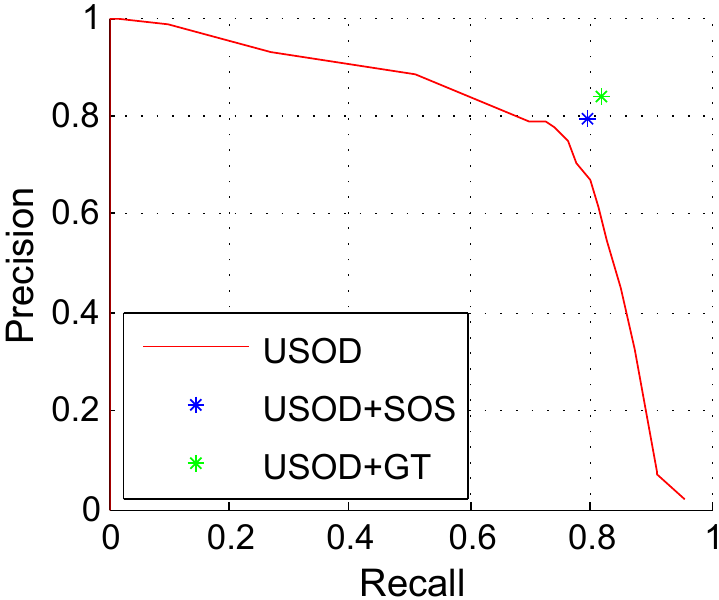}
    \caption{Precision-Recall curve of USOD, and the performance of USOD+SOS and USOD+GT.}\label{fig:sod}
\end{figure}

\textbf{Results.} The results are reported in Table~\ref{tab:sod}. Fig.~\ref{fig:sod} shows the PR curve of USOD compared to the precision and recall rates of USOD+SOS and USOD+GT. As we can see, USOD+SOS significantly outperforms the baseline USOD, obtaining an absolute increase of about 4\% in F-measure. This validates the benefit of adaptively tuning the PR parameter based on the SOS model. When the groundtruth number of objects is used (USOD+GT), another absolute increase of 3\% can be attained, which is the upper bound for the performance improvement. Table~\ref{tab:sod} also reports the performance of each method on images with salient objects. On this subset of images, using SOS improves the baseline USOD by about 1 absolute percentage point in F-measure. This suggests that our CNN SOS model is not only helpful for suppressing detections on background images, but is also beneficial by determining the number of detection windows for images with salient object.

\begin{table}
\caption{Salient object detection performance on the MSO dataset. For the baseline USOD, we report its performance using the PR parameter that gives the optimal F-measure (\%). We also report the performance of each method on a subset of the MSO dataset, which only contain images with salient objects (see Obj.\ Img.\ below).}
\label{tab:sod}
\centering
\begin{tabular}{ccccc}
\toprule
            & & Prec. & Rec. & F-score \\\cmidrule{2-5}
    \multirow{ 3}{*}{Full Dataset} &USOD  & 77.5     & 74.0     &75.7\\
                                   &USOD+SOS      & 79.6     & 79.5     & 79.5\\
                                   &USOD+GT       & 83.9     & 81.7     & 82.8 \\\cmidrule{2-5}
    \multirow{ 3}{*}{Obj. Img.} &USOD  & 78.0     & 81.0     & 79.4\\
                                   &USOD+SOS      & 79.5     & 81.8     & 80.6\\
                                   &USOD+GT       & 83.9     & 81.7     & 82.8 \\
   \bottomrule
\end{tabular}
\end{table}

%\begin{table}
%\caption{Subtizing for salient object detection. The baseline is the object detection method from \citep{zhang2015SOD}. We report the operation point with the optimal F-score (\%) for the baseline.}
%\label{tab:sod}
%\centering
%\begin{tabular}{ccccc}
%\toprule
%                          & & Precision & Recall & F-score \\\cmidrule{2-5}
%   \multirow{ 3}{*}{USOD} & baseline  & 77.5     & 74.0     &75.7\\
%                          & +SBT      & 79.6     & 79.5     & 79.5\\
%                          & +GT       & 83.9     & 81.7     & 82.8 \\\cmidrule{2-5}
%   \multirow{ 3}{*}{MBox} & baseline  & 65.9     & 62.9     & 64.3\\
%                          & +SBT      & 76.6     & 73.2     & 74.9\\
%                          & +GT       & 78.6     & 74.4     & 76.5 \\
%   \bottomrule
%\end{tabular}
%\end{table}

\textbf{Cross-dataset generalization test for identifying background images.} Detecting background images is also useful for tasks like salient region detection and image thumbnailing \citep{wang2012salient}. To test how well the performance of our SOS model generalizes to a different dataset for detecting the presence of salient objects in images, we evaluate it on the web thumbnail image test set proposed by \cite{wang2012salient}. The test set used by \cite{wang2012salient} is composed of 5000 thumbnail images from the Web, and 3000 images sampled from the MSRA-B \cite{liu2011learning} dataset. $50\%$ of these images contain a single salient object, and the rest contain no salient object. Images for MSRA-B are resized to  $130\times 130$ to simulate thumbnail images \citep{wang2012salient}.

In Table~\ref{tab:sod2}, we report the detection accuracy of our CNN SOS model, in comparison with the 5-fold cross-validation accuracy of the best model reported by \cite{wang2012salient}. Note that our SOS model is trained on a different dataset, while the compared model is trained on a subset of the tested dataset via cross validation. Our method outperforms the model of \cite{wang2012salient}, and it can give fast prediction without resorting to any salient object detection methods. In contrast, the model of \cite{wang2012salient} requires computing several saliency maps, which takes over 4 seconds per image as reported by \cite{wang2012salient}.

\begin{table}
\caption{ Recognition accuracy in predicting the presence of salient objects on the thumbnail image dataset \citep{wang2012salient}. We show the 5-fold cross validation accuracy reported in \citep{wang2012salient}. While our method is trained on the MSO dataset, it generalizes well to this other dataset.}
\label{tab:sod2}
\centering
\begin{tabular}{ccc}
\toprule
                 & \cite{wang2012salient} & Ours \\\cmidrule{1-3}
   accuracy (\%) & 82.8 & 84.2\\
\bottomrule
\end{tabular}
\end{table}

\subsection{Image Retrieval}

In this section, we show an application of SOS in Content Based Image Retrieval (CBIR). In CBIR, many search queries refer to object categories. It is useful in many scenarios that users can specify the number of object instances in the retrieved images. For example, a designer may search for stock images that contain two animals to illustrate an article about couple relationships, and a parent may want to search his/her photo library for photos of his/her baby by itself.

We design an experiment to demonstrate how our SOS model can be used to facilitate the image retrieval for number-object (\emph{e.g.}\ ``three animals") search queries. For this purpose, we implement a tag prediction system. Given an image, the system will output a set of tags with confidence scores. Once all images in a database are indexed using the predicted tags and scores, retrieval can be carried out by sorting the images according to the confidence scores of the query tags.

\textbf{The tag prediction system.} Our tag prediction system uses 6M training images from the Adobe Stock Image website. Each training image has 30-50 user provided tags. We pick
about 18K most frequent tags for our dictionary. In practice, we only keep the first 5 tags for an image as we empirically find that first few tags are usually more relevant. Noun Tags and their plurals are merged (\emph{e.g.}\ ``person" and ``people" are treated as the same tag). We use a simple KNN-base voting scheme to predict image tags. Given a test image and a Euclidean feature space, we retrieve the 75 nearest neighbors in our training set using the distance encoded product quantization scheme of \cite{Heo_2014_CVPR}. The proportion of the nearest neighbors that have a specific tag is output as the tag's confidence score. The Euclidean feature space for the KNN system is learned by a CNN model. We use the GoogleNet architecture and use the last 1024D average pooling layer as our feature space. Details about the CNN feature embedding training are included in the supplementary material.
\begin{figure*}
\centering
	\includegraphics[width = 1\linewidth]{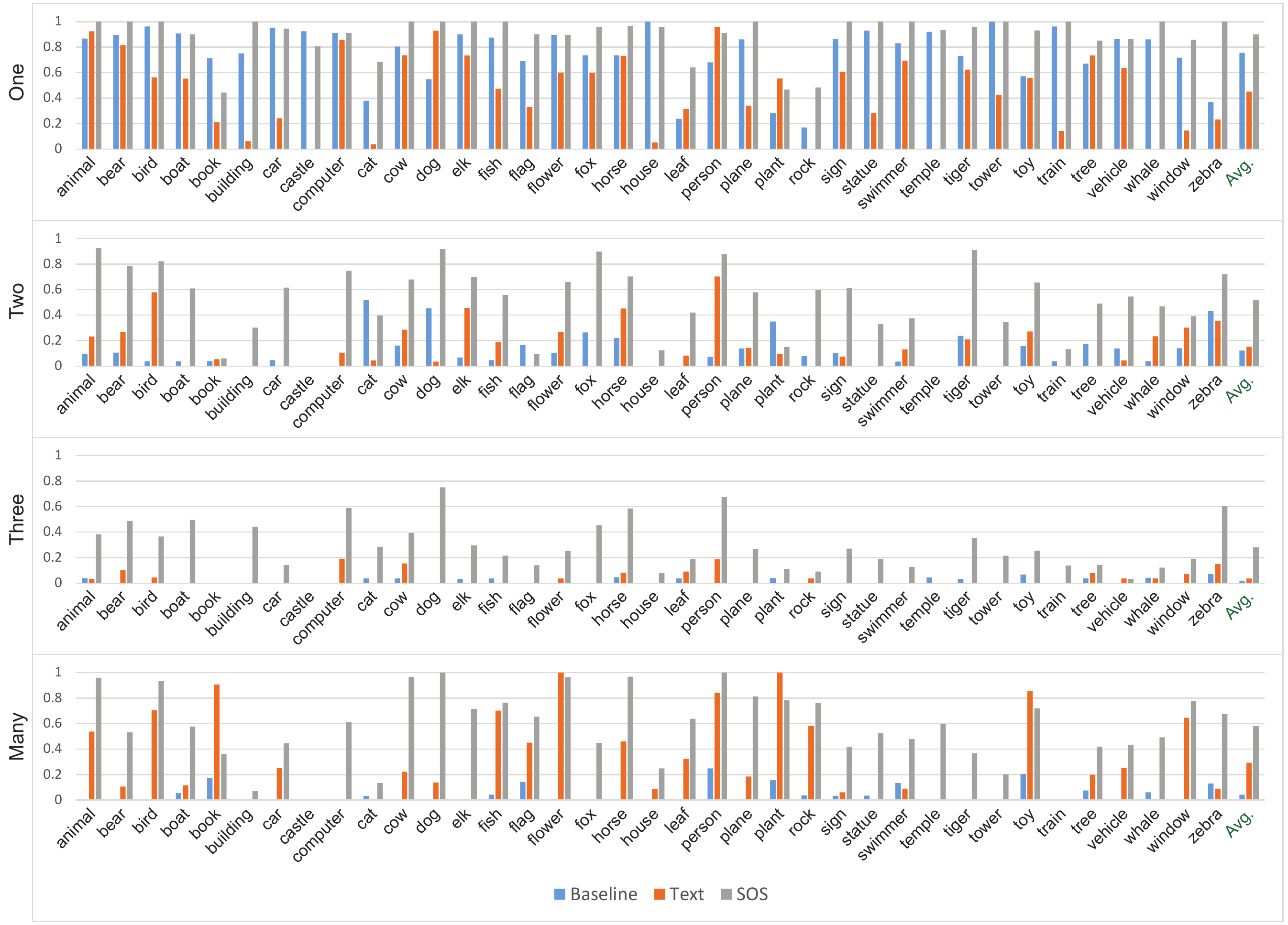}
    \caption{nDCG scores for compared methods. For each object class, we use different methods to retrieve images of one/two/three/many object(s) of such class. The last column shows the average nDCG scores across different object classes.}\label{fig:ndcg}
\end{figure*}
\begin{figure*}
\centering
	\includegraphics[width = 1\linewidth]{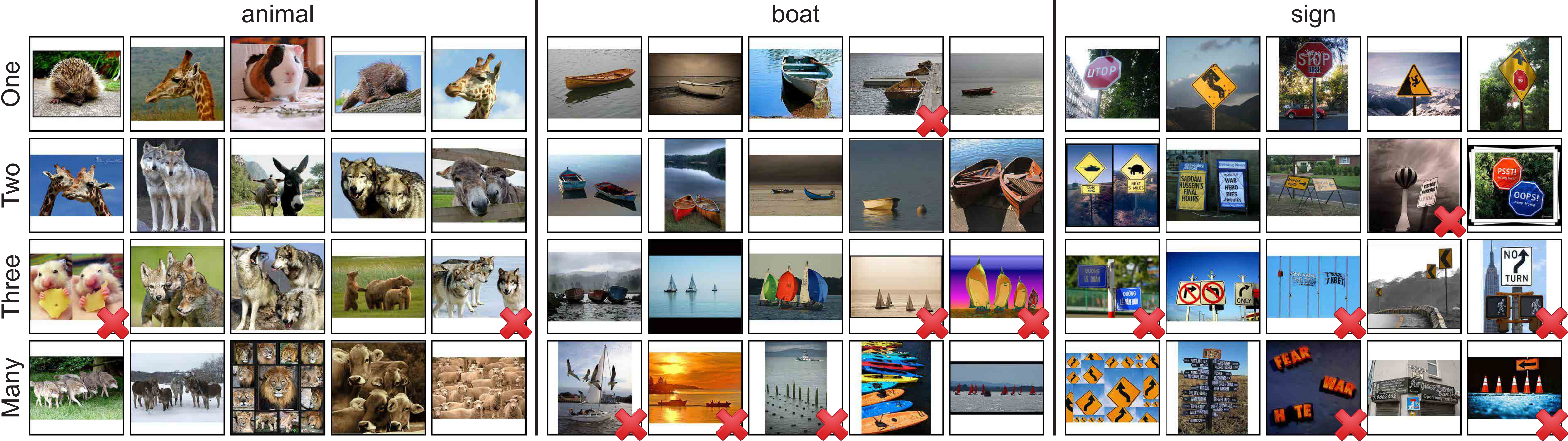}
    \caption{Sample results of the SOS-based method for number-object image retrieval. The base object tags are shown above each block. Each row shows the top five images for a number group (one/two/three/many). Irrelevant images are marked by a red cross.}\label{fig:nus_sample}
\end{figure*}

\textbf{Dataset.} We use the public available NUS-WIDE dataset as our test set \citep{chua2009nus}, which contains about 270K images. We index all the images of NUS-WIDE using our tag prediction system for all the tags of our dictionary. The NUS-WIDE dataset has the annotation of 81 concepts, among which we pick all the concepts that correspond to countable object categories as our base test queries (see Fig.~\ref{fig:ndcg} for the 37 chosen concepts). For a base test query, say ``animal", we apply different test methods to retrieve images for four sub-queries, ``one animal", ``two animals", ``three animals" and ``many animals", respectively. Then all the retrieved images for ``animal" by different test methods are mixed together for annotation. We ask three subjects to label each retrieved image as one of the four sub-queries or none of the sub-queries (namely a five-way classification task). The subjects have no idea which test method retrieved which image. Finally, the annotations are consolidated by majority vote to produce the ground truth for evaluation.

\textbf{Methods.} Given the tag confidence scores of each image by our tag prediction system, we use different methods to retrieve images for the number-object queries.
\begin{itemize}
  \item Baseline. The baseline method ignores the number part of a query, and retrieves images using only the object tag.
  \item Text-based method. This method treats each sub-query as the combination of two normal tags. Note that both the object tags and the number tags are included in our dictionary. We multiply the confidence scores of the object tag with the confidence scores of the number tag (``one", ``two", ``three" or ``many"). Then the top images are retrieved according to the multiplied scores.
  \item SOS-based method. This method differs from the text-based method in that it replaces the number tag confidence score with the corresponding SOS confidence score. For a number tag ``one/two/three/many", we use the SOS confidence score for 1/2/3/4+ salient object(s).
  %For a sub-query, say ``one/two/three/many animal(s)", we multiply each image's tag confidence score of ``animal" with the corresponding SOS confidence scores for 1/2/3/4+ salient object(s).
\end{itemize}

\textbf{Evaluation Metric.} The widely used Average Precision (AP) requires annotation of the whole dataset for each number-object pair, which is too expensive. Therefore, we use the normalized Discounted Cumulative Gain (nDCG) metric, which only looks at the top retrieved results. The nDCG is used in a recent image retrieval survey paper by \cite{li2015socializing} for benchmarking various image retrieval methods. The nDCG is formulated as
\begin{equation}
   {nDCG}_h(t) = \frac{{DCG}_h(t)}{{IDCG}_h(t)},
\end{equation}
where $t$ is the test query, $DCG_h(t) = \sum_{i=1}^h \frac{2^{rel_i}-1}{\log_2(i+1)}$, and  $rel_i$ denotes the tag relevance of the retrieved image at position $i$. In our case, $rel_i$ is either 0 or 1. The $IDCG_h(t)$ is the maximum possible $DCG$ up to position $h$. We retrieve 20 images for each method, so we set $h=20$ and assume that there are at least 20 relevant images for each query.

\textbf{Results.} The nDCG scores of our SOS-based method, the text-based method and the baseline method are reported in Fig.~\ref{fig:ndcg}. The SOS-based method gives consistently better average nDCG scores across queries for different numbers of objects, especially for the queries for more than one object. The scores of the SOS-based method for the group ``three" are overall much lower than for the other groups. This is because the accuracy of our SOS is relatively lower for three objects. Moreover, there are many object categories that lack images with three objects, \emph{e.g.}\ ``statue", ``rock", \emph{etc.}

The baseline method gives pretty good nDCG scores for a single object, but for the other number groups, its performance is the worst. This reflects that images retrieved by a single object tag tend to contain only one dominant object. Note that it is often favorable that the retrieved images present a single dominant object of the searched category when no number is specified. When using SOS, the performance in retrieving images of one object is further improved, indicating it can be beneficial to apply SOS by default for object queries.

The text-based method is significantly worse than our SOS-based method across all number groups. We observe that when a query has a number tag like ``one", ``two" and ``three", the retrieved images by the text-based method tends to contain the given number of people. We believe that this is because these number tags often refer to the number of people in our training images. This kind of data bias obstructs the simple text-based approach to handling number-object queries. In contrast, our SOS-based method can successfully retrieve images for a variety of number-object queries thanks to the category agnostic nature of our SOS formulation. Sample results of our SOS-based method are shown in Fig.~\ref{fig:nus_sample}.

\section{Conclusion}

In this work, we formulate the Salient Object Subitizing (SOS) problem, which aims to predict the existence and the number of salient objects in an image using global image features, without resorting to any localization process. We collect an SOS image dataset, and present a Convolutional Neural Network (CNN) model for this task. We leverage simple synthetic images to improve the CNN model training. Extensive experiments are conducted to show the effectiveness and generalizability of our CNN-based SOS method. We visualize that the features learned by our CNN model capture generic visual patterns that are useful for subitizing, and show how our model can generalize to unseen object categories. The usefulness of SOS is demonstrated in unconstrained salient object detection and content-based image retrieval. We show that our SOS model can improve the state-of-the-art salient object detection method, and it provides an effective solution to retrieving images by number-object queries.

\section*{Acknowledgments}
This research was supported in part by US NSF grants 0910908 and 1029430, and gifts from Adobe and NVIDIA.

\bibliographystyle{spbasic}      % basic style, author-year citations
\small
\bibliography{egbib}

\end{document}